\documentclass[pdflatex,sn-vancouver-ay]{sn-jnl}

\usepackage{graphicx}%
\usepackage{multirow}%
\usepackage{amsmath,amssymb,amsfonts}%
\usepackage{amsthm}%
\usepackage{mathrsfs}%
\usepackage[title]{appendix}%
\usepackage{xcolor}%
\usepackage{textcomp}%
\usepackage{manyfoot}%
\usepackage{booktabs}%
\usepackage{algorithm}%
\usepackage{algorithmicx}%
\usepackage{algpseudocode}%
\usepackage{listings}%
\usepackage{xcolor}
\usepackage{afterpage}
\usepackage{rotating}
\usepackage{booktabs}
\usepackage{multirow}
\usepackage{makecell}
\usepackage{booktabs}
\usepackage{graphicx}
\usepackage{float}
\usepackage{placeins}

\theoremstyle{thmstyleone}%
\theoremstyle{thmstyletwo}%

\theoremstyle{thmstylethree}%

\begin{document}

\title[Article Title]{CAMO: A Class-Aware Minority-Optimized Ensemble
for Robust Language Model Evaluation on Imbalanced Data}


\author*[1]{\fnm{Mohamed} \sur{Ehab}}\email{me338484@gmail.com}

\author[1]{\fnm{Ali} \sur{Hamdi}}\email{ahamdi@msa.edu.eg}

\author[2]{\fnm{Khaled} \sur{Shaban}}\email{khaled.shaban@qu.edu.qa}

\affil*[1]{\orgdiv{Faculty of computer science}, \orgname{October University for Modern Science \& Arts}, \city{Giza},\country{Egypt}}

\affil[2]{\orgdiv{Department of Computer Science and Engineering}, \orgname{Qatar University}, \city{Doha}, \country{Qatar}}


\abstract{Real-world categorization is severely hampered by class imbalance because traditional ensembles favor majority classes, which lowers minority performance and overall F1-score. We provide a unique ensemble technique for imbalanced problems called CAMO (Class-Aware Minority-Optimized). Through a hierarchical procedure that incorporates vote distributions, confidence calibration, and inter-model uncertainty, CAMO dynamically boosts underrepresented classes while preserving and amplifying minority forecasts.We verify CAMO on two highly unbalanced, domain-specific benchmarks: the DIAR-AI/Emotion dataset and the ternary BEA 2025 dataset. We benchmark against seven proven ensemble algorithms using eight different language models (three LLMs and five SLMs) under zero-shot and fine-tuned settings.With refined models, CAMO consistently earns the greatest strict macro F1-score, setting a new benchmark. Its benefit works in concert with model adaptation, showing that the best ensemble choice depends on model properties.This proves that CAMO is a reliable, domain-neutral framework for unbalanced categorization.}

\keywords{Ensemble Learning, Class Imbalance, Minority Class Optimization, CAMO, Imbalanced Classification, Macro F1-Score, Language Model Evaluation}



\maketitle

\section{Introduction}\label{sec1}

Class imbalance is a fundamental challenge in areas of machine learning where minority groups are important, yet they remain largely ignored. Due to this imbalance, traditional algorithms focus on the majority classes, which, in turn, places the macro F1 score (and other macro-level metrics) at a significant risk \cite{daheim2024stepwiseverificationremediationstudent}. Standard ensemble methods like majority voting and confidence-weighted averaging make the situation even worse by marginalizing the minority class contributions even more.This creates a significant methodological gap because ensemble methods focused on maintaining and improving predictions on minority classes are still in the very early stages of development. This is particularly true in fields that require complex and advanced multi-class classification, such as educational AI, where the comments of teachers are categorized into three classes, including the statistically infrequent yet pedagogically important “To some Extent” class. Here, we are faced with problems on a much larger scale where the available technology and systems are lacking the advanced reasoning required\cite{hendrycks2021measuringmassivemultitasklanguage, hendrycks2021measuringmathematicalproblemsolving}. The emotional states of minorities are even more neglected in the area in question. Emotion recognition depends on accuracy of the performance. The arrival of AI tutors (for example, AutoTutor) \cite{article} offers promise of personalized education, but this kind of technology will need much more than conventional metrics of accuracy to gauge the complexity of recognition of the emotion. In different contexts to solve the class imbalance issue, we propose the new ensemble method CAMO (Class-Aware Minority-Optimized). CAMO is based on the view that the predictions of the minority class are different and are not to be thrown away as noise, but rather and signal that should be enhanced and retained. The methodology outlines a multi-dimensional hierarchical decision system which includes, but is not limited to, distribution of votes, adjustment of confidence, model uncertainty, and thresholds merging.
When used together with other methods, and with appropriate conditions, CAMO improves the prediction of the minority classes and thus, improves the score of the macro F1. This is apparent in varying situations. With this paper, we provide an all-encompassing analysis from the viewpoint of two imbalanced and ternary structured dense datasets - BEA 2025 Mistake Identification Track, and DIAR-AI/Emotion dataset. The experimental setup includes the analysis of eight language models - three being large language models and the other five being small language models - under zero-shot and fine-tuned conditions.
We create one of the most comprehensive comparative evaluations for imbalanced classification to date by benchmarking CAMO against seven well-known ensemble algorithms.We provide a novel domain-agnostic ensemble technique with explicit minority class preservation features called \textbf{CAMO}. \textbf{Comprehensive validation} on two different imbalanced datasets, showing improved performance with optimized models. The efficacy of CAMO is dependent on task adaptability and model calibration, according to contingency analysis. Comprehensive benchmarking across many architectures, providing useful advice for unbalanced categorization.we establish CAMO as a strong framework for resolving class imbalance that can be applied to any classification assignment where minority classes are underrepresented but essential for fair performance.

\section{Related Work}
\label{sec:related_work}

\subsection{Ensemble Methods and Class Imbalance}
Minority predictions are marginalized by ensemble techniques like majority voting and confidence-weighted averaging, which frequently favor majority classes. Ambiguous minority scenarios continue to be a challenge for recent transformer-based ensembles \cite{saha-etal-2025-nlip}. While pedagogically-informed reasoning approaches \cite{roh-bang-2025-bea} are still vulnerable to prompt design, disagreement-aware strategies \cite{hikal-etal-2025-msa} have been proposed to maintain minority votes in circumstances of model disagreement. These initiatives demonstrate the necessity of stronger minority-aware ensemble designs.

\subsection{Evaluation in Specialized Imbalanced Domains}
Evaluation frameworks that are specific to some domains can show us some gaps between the AI systems and human judgments \cite{tack2022aiteachertestmeasuring}, and these gaps require more than just measurements of accuracy. MRBench \cite{kochmar-etal-2025-findings} benchmarks class imbalance within some middle ranges of the educational assessment. Emotion classification datasets are quite lacking in some specific emotions \cite{kermani-etal-2025-systematic,henrichsen2025twostagereasoninginfusedlearningimproving}. Frameworks that capture the need for more balanced systems show us the interactive measurement frameworks \cite{demszky2021measuringconversationaluptakecase} and the frameworks that address the uncertainties \cite{wang2024bridgingnoviceexpertgapmodels}.

\subsection{Handling Imbalance and Minority Classes}
Automated assessment has also been complicated by the well-known class imbalance issue \cite{hendrycks2021measuringmathematicalproblemsolving}, where the benchmarks in majority classes are 79\% and 7\% for the minority classes \cite{saha-etal-2025-nlip}. Simple approaches like weighted loss and oversampling tend to underperform in capturing the intricacies of the minority classes. Deep Ensembles \cite{lakshminarayanan2017simplescalablepredictiveuncertainty} and uncertainty-based approaches are suggesting models to manage uncertainty and avoid overestimating. While preference-aligned models \cite{siddiqui2025self} may enhance interpretability, they still struggle with the underrepresented classes, and studies on fine-tuning, prompting, and RAG \cite{kermani-etal-2025-systematic} identify cost and minority class performance trade-offs.

\subsection{Reasoning-Augmented and Efficient Adaptation}
Strong, open foundation models such as LLaMA \cite{touvron2023llamaopenefficientfoundation} and LlaMa 2 \cite{touvron2023llama2openfoundation} support the trend toward efficient specialization.
Common courses benefit much from reasoning-infused learning \cite{henrichsen2025twostagereasoninginfusedlearningimproving}, but minority classes suffer greatly from it. Although they rely on expensive preference data, preference optimization techniques \cite{siddiqui2025self} improve the quality of explanations. Task-specific tuning with limited parameters \cite{11364256} is made possible by efficient adaptation via PEFT methods such as LoRA \cite{hu2021loralowrankadaptationlarge} and QLoRA \cite{dettmers2023qloraefficientfinetuningquantized}, which have been successfully applied in domains such as mental health classification \cite{kermani-etal-2025-systematic} and pedagogical evaluation \cite{roh-bang-2025-bea}.

\subsection{Research Gaps and Our Contribution}
Minority performance is sacrificed for overall accuracy in current ensemble approaches. Systematic comparisons lack minority-specific ensemble methods \cite{kermani-etal-2025-systematic}, and reasoning-augmented approaches are vulnerable to imbalance \cite{henrichsen2025twostagereasoninginfusedlearningimproving,siddiqui2025self}. In order to close these gaps, CAMO introduces: (1) multi-stage decision processes that incorporate voting, thresholds, and uncertainty; (2) dynamic minority boosting via confidence and uncertainty; and (3) thorough benchmarking across architectures and training paradigms. In contrast to earlier research, CAMO incorporates uncertainty-aware adaptation to guarantee fair performance in every class.

\section{Methodology}
\label{sec:methodology}
In order to check how CAMO does when dealing with classification tasks where some classes have much more data than others, a complicated series of experiments was set up. The process – including getting the data ready, picking which models to use, adjusting those models, building a group of them, and thoroughly testing everything – is in the plan to make certain the results can be duplicated, and will work well with different kinds of data.

\subsection{Datasets and Preprocessing}
To test CAMO’s capacity to handle different types of data skew, we used two ternary classification datasets with seriously small minority classes: firstly, the BEA 2025 Mistake Identification Track \cite{kochmar-etal-2025-findings} – this has a minority class, “To some extent”, of around 7\%; and secondly, the DIAR-AI/Emotion dataset, which shows significant imbalance in its particular emotion groups. Both of these show long-tail distributions. BEA employs time-based splitting (70\%/30\%) to keep things in order, though DIAR-AI uses established divisions. To make sure all models and training were comparable, the data going in was made into prompts which had a clear structure.

\subsection{Model Selection and Fine-Tuning}
We use eight different language models for BEA—including SLMs like Phi-3-mini-4k-instruct \cite{abdin2024phi-}, DeepSeek-R1-1.5B \cite{DeepSeekAI2025DeepSeekR1},Qwen3-0.6B \cite{Yang2025Qwen3TR}, Llama-3.2-1B-Instruct, and Falcon3-1B-Instruct \cite{almazrouei2023falconseriesopenlanguage} to test CAMO across architectural scales and training conditions, LLMs like LLama-3.1-8B and Mistral-7B\cite{jiang2023mistral7b},DeepSeek-R1-Distill-Llama-8B. Llama-3.1-8B and Llama-3.2-1B-Instruct \cite{grattafiori2024llama3herdmodels} with multi-seed training are used for DIAR-AI. With domain-specific hyperparameter tuning, gradient checkpointing, and 4-bit quantization for memory efficiency, LoRA \cite{hu2021loralowrankadaptationlarge,11364256} is used to fine-tune all models for parameter-efficient adaptation. In addition to lowering variance, multi-seed training yields uncertainty estimates.

\subsection{Ensemble Strategies}
We compare CAMO to well-established ensemble baselines: for BEA, seven techniques, such as majority voting and confidence-weighted voting; for DIAR-AI, modern methods like two-stage reasoning-infused learning \cite{henrichsen2025twostagereasoninginfusedlearningimproving}, systematic LLM evaluation \cite{kermani-etal-2025-systematic}, and self-explaining emotion classification \cite{siddiqui2025self}. Inspired by verification-based reasoning techniques \cite{cobbe2021trainingverifierssolvemath}, CAMO employs a hierarchical decision procedure that incorporates unanimity checks, minority signal detection, confidence calibration, uncertainty-based minority preference, and dynamic boosting. In a variety of model disagreement and confidence dispersion scenarios, this structured yet flexible technique is intended to maintain and enhance minority class predictions.

\begin{figure}[H]
    \makebox[\textwidth][c]{
        \includegraphics[width=0.95\paperwidth]{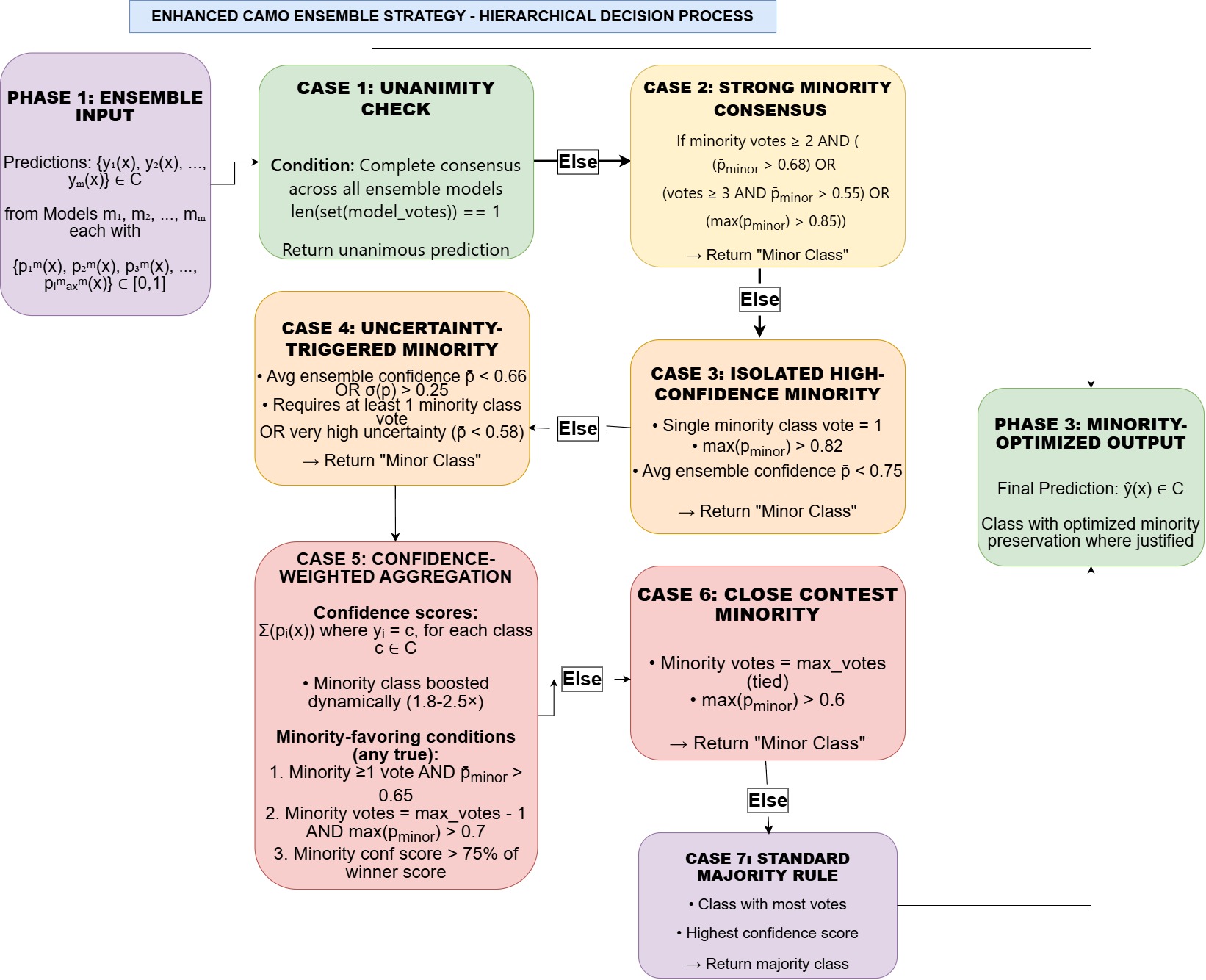} 
    }
    \caption{
        CAMO ensemble decision process architecture at a high level.
        In order to optimize for minority class preservation, the hierarchical pipeline moves from unanimity checks to minority-aware boosting, combining vote distribution, confidence calibration, and inter-model uncertainty.
    }
    \label{fig:camodesign}
\end{figure}

\subsection{CAMO: Mathematical Formulation}
\label{subsec:camo_math}

We provide a formal description of the CAMO ensemble method, which combines dynamic class-specific boosting, uncertainty-aware decision boundaries, and hierarchical minority-signal detection.

Let a set of classes be used to define the categorization task.
\[
\mathcal{C} = \{c_1, \dots, c_K\},
\]
and an ensemble of models
\[
\mathcal{M} = \{m_1, \dots, m_M\}.
\]
Minority classes are represented by the subset $\mathcal{C}_{\mathrm{min}} \subseteq \mathcal{C}$, while majority classes are represented by $\mathcal{C}_{\mathrm{maj}} = \mathcal{C} \setminus \mathcal{C}_{\mathrm{min}}$.

Each model $m\in\mathcal{M}$ generates a projected label for input $x$.
\[
\hat{y}_m(x)\in\mathcal{C}
\]
with confidence score
\[
p_m(x)\in[0,1].
\]

\subsubsection{Aggregate Voting and Confidence Statistics}

For every class $c\in\mathcal{C}$, the ensemble vote count is defined as
\[
V(c) = \sum_{m\in\mathcal{M}} \mathbb{I}[\hat{y}_m(x)=c].
\]

The dispersion of the mean ensemble confidence is provided by
\[
\bar{p} = \frac{1}{M}\sum_{m\in\mathcal{M}} p_m(x), 
\qquad
\sigma_p = \sqrt{\frac{1}{M}\sum_{m\in\mathcal{M}} (p_m(x)-\bar{p})^2}.
\]

For each minority class $c\in\mathcal{C}_{\mathrm{min}}$ with $V(c)>0$,
\[
\bar{p}_c = \frac{1}{V(c)}\!\!\sum_{m:\hat{y}_m=c} p_m(x), 
\qquad
p_c^{\max} = \max_{m:\hat{y}_m=c} p_m(x),
\]
and if $V(c)=0$, both amounts are zero.

\subsubsection{Hierarchical Decision Structure}

CAMO uses a seven-stage hierarchical review to forecast the final class.  
Define
\[
\mathcal{D}(x) = \operatorname*{argpriority}_{i\in\{1,\dots,7\}} \mathcal{C}_i(x),
\]
where the final output is determined by the first satisfied condition.

\paragraph{1. Unanimity}
\[
\mathcal{C}_1:\quad 
\exists c\in\mathcal{C}\ \text{s.t.}\ V(c)=M.
\]

\paragraph{2. Strong Minority Consensus}
{\footnotesize
For each $c\in\mathcal{C}_{\mathrm{min}}$,
\[
\mathcal{C}_2^c:\;
V(c)\ge\theta_1^c
\land
\big(
\bar{p}_c>\tau_1^c
\lor
[V(c)\ge\theta_2^c \land \bar{p}_c>\tau_2^c]
\lor
p_c^{\max}>\tau_3^c
\big).
\]
}

\paragraph{3. Isolated High-Confidence Minority Vote}
\[
\mathcal{C}_3^c:\;
V(c)=1 \ \land\ p_c^{\max}>\tau_4^c \ \land\ \bar{p}<\tau_5^c.
\]

\paragraph{4. Uncertainty-Triggered Minority Prioritization}
\[
\mathcal{C}_4:\;
(\bar{p}<\tau_6 \ \lor\ \sigma_p>\tau_7)
\ \land\
\bigvee_{c\in\mathcal{C}_{\mathrm{min}}}
\left( V(c)\ge 1 \ \lor\ \bar{p}<\tau_8^c \right).
\]

\paragraph{5. Confidence-Weighted Aggregation with Boosting}

The class score is
\[
S(c) = \sum_{m:\hat{y}_m=c} p_m(x)\,\beta(c,V(c),\bar{p}),
\]
with boost function
\[
\beta(c,V(c),\bar{p})=
\begin{cases}
B_c(V(c),\bar{p}), & c\in\mathcal{C}_{\mathrm{min}},\\[4pt]
1, & c\in\mathcal{C}_{\mathrm{maj}}.
\end{cases}
\]

\paragraph{6. Minority Vote Under Competitive Dominance}
\[
\mathcal{C}_6:\;
\exists c\in\mathcal{C}_{\mathrm{min}}:
\ \ V(c)=\max_{c'\in\mathcal{C}} V(c')
\ \land\ 
p_c^{\max}>\tau_9^c.
\]

\paragraph{7. Standard Majority Rule}
If none of the previous conditions apply,
\[
\mathcal{C}_7:\quad 
\mathcal{D}(x) = \arg\max_{c\in\mathcal{C}} S(c).
\]

\subsubsection{Dynamic Minority Boost Function}

For each minority class $c\in\mathcal{C}_{\mathrm{min}}$, the boosting term is defined as
\[
B_c(v,\bar{p}) =
\min\left(
\beta_c^{\mathrm{base}}
+
\sum_{i=1}^{4} \alpha_i^c\,\mathbb{I}[\mathrm{condition}_i],
\;
\beta_c^{\max}
\right),
\]
where the binary conditions correspond to the following triggers:
\[
v\ge 2,\qquad
v\ge 3,\qquad
\bar{p}<0.7,\qquad
\bar{p}<0.6.
\]

\subsection{Evaluation Framework}
\subsubsection{Comprehensive Evaluation Metrics}
Strict macro F1-score is the major statistic for BEA; accuracy and lenient macro F1 are the additional metrics. Macro-averaged precision, recall, and F1-score for emotion, with a focus on minority classes (love, surprise). Per-class analysis, weighted F1, and fairness measures (F1 gap between majority/minority classes) are also included in both.

\subsubsection{Statistical Validation and Reproducibility}
We used version-controlled settings, fixed random seeds, statistical significance testing, confidence intervals through resampling, and thorough logging. Efficient experimentation was assured while preserving rigorous, repeatable evaluation across architectures and training paradigms thanks to modern deep learning frameworks and improved infrastructure.

\section{Experimental Setup and Implementation}

\subsection{Parameter Settings and Hyperparameters}

\subsubsection{Dataset-Specific Training Configurations}
To guarantee optimal adaptation while maintaining fair comparison within each domain, we kept different but consistently applied hyperparameter configurations for every dataset. For memory efficiency, all models used the same 4-bit quantization. For the \textbf{BEA Dataset Configuration}, we used the AdamW optimizer with torch implementation, enabled gradient checkpointing, set LoRA rank to 32, set the maximum sequence length to 2,048 tokens, a global batch size of 16 (with micro batch size 4 and gradient accumulation steps 4), a learning rate of $5 \times 10^{-6}$ with cosine scheduling, weight decay of 0.01, maximum gradient norm of 0.3, six epochs, and 100 warmup steps. Training seeds [42, 43, 44, 45, 46] were utilized for the multi-seed ensemble, LoRA alpha was set to 64, LoRA dropout was set to 0.05, and targeted modules \texttt{["qkv proj", "o proj", "gate up proj", "down proj"]}. We used the AdamW 8-bit optimizer, enabled gradient checkpointing, set early stopping patience to 3 epochs (based on macro F1-score), set LoRA rank to 32, LoRA alpha to 64, LoRA dropout to 0.05, a maximum sequence length of 256 tokens, an effective batch size of 64 (with micro batch size 16 and gradient accumulation steps 4), a learning rate of $2 \times 10^{-4}$ with linear scheduling. 

\subsubsection{Controlled Experimental Factors}
We strictly controlled the following variables within each dataset to guarantee thorough and repeatable comparisons: Fixed variables include train\newline/validation/test splits specific to each dataset, uniform input formatting and prompt templates within each domain, the same evaluation metrics and calculation techniques for each dataset, the same computational budget for each model category, fixed random seeds for all stochastic operations, and uniform 4-bit quantization for all models.

\textbf{Systematic Variations:} \begin{itemize} \item \textbf{BEA Dataset:} Model architecture (eight models), training condition (fine-tuned vs. baseline), ensemble strategy (eight methods), model scale category (SLM vs. LLM) \item \textbf{Emotion Dataset:} Minority class boosting techniques, model uncertainty quantification, and training seed variation (five distinct seeds) \end{itemize}

\subsubsection{Ensemble Strategy Implementation}
Every ensemble technique was put into practice as a separate module with uniform interfaces. We used the following baseline techniques for the BEA dataset:

\begin{table}[h!]
\centering
\small
\begin{tabular}{lp{10cm}}
\hline
\multicolumn{2}{c}{\textbf{BEA Baseline Ensemble Strategies}} \\
\hline
\textbf{Method} & \textbf{Description} \\
\hline
\textbf{Majority Voting} & Simple plurality-based aggregation \\
\textbf{Confidence-Weighted} & Softmax probability weighted voting \\
\textbf{Class-Balanced} & Inverse class frequency weighting \\
\textbf{Dynamic Threshold} & Adaptive decision boundaries \\
\textbf{Uncertainty-Aware} & Entropy-based model weighting \\
\textbf{Meta-Ensemble} & Stacked generalization approach \\
\textbf{MSA Paper's Baseline} & Reference implementation from prior ensemble research \\
\hline
\end{tabular}
\end{table}
\FloatBarrier
\textbf{CAMO Strategy Implementation:}
The CAMO ensemble uses an upgraded hierarchical decision process with settings tailored for each dataset: \textbf{Unanimity threshold:} 100\% agreement,\textbf{Strong minority detection:} $\geq 2$ votes with confidence validation,\textbf{High-confidence minority detection:} Single vote $> 0.82$ confidence (BEA), $> 0.75$ for "surprise" and $> 0.82$ for "love" (Emotion) , \textbf{Uncertainty threshold:} Average confidence $< 0.66$ (both datasets) ,\textbf{Dynamic boost range:} 1.5--2.5 (adaptive, with class-specific modifications).CAMO uses class-specific parameters to improve minority class handling for the Emotion dataset: \textbf{Minority classes:} "surprise" ($\sim$3\%) and "love" ($\sim$8\%),{Class-specific boosting:} "Surprise" (2.5) has a greater base increase than "love" (1.8),{Improved decision tree:} Seven decision-making techniques with a guaranteed backup plan, \textbf{Surprise prioritization:} More permissive surprise detection levels,{Dynamic boosting:} Context-aware minority preference rises with uncertainty and the number of votes.While preserving methodological consistency and reproducibility, our dual-dataset experimental approach enables thorough assessment of CAMO's efficacy across various imbalance circumstances, model topologies, and minority class characteristics.

\section{Results}
\label{sec:results}

This section provides a thorough assessment of the suggested CAMO ensemble technique in two separate imbalanced classification domains using several experimental setups suitable for the features and assessment goals of each dataset.

\subsection{BEA 2025 Dataset Results}

Under both fine-tuned and baseline (zero-shot) settings, we compare CAMO against seven baseline ensemble methods across eight language models (three large language models and five minor language models). The analysis covers rigorous and lenient macro F1-scores, per-model comparisons across architectural families, statistical significance testing, and ablation studies isolating the contribution of CAMO's fundamental components. Further evaluations concentrate on computational efficiency trade-offs, performance on the crucial "To some extent" minority class, and qualitative case studies that demonstrate CAMO's decision-making process in educational assessment situations.

\subsection{DIAR-AI/Emotion Dataset Results}

To work out how good CAMO is at classifying emotion, we used the Llama-3.1-8B model – trained using five separate random seeds in what’s called a multi-seed ensemble – for the emotion categorisation job. The experiment looks at CAMO, in both its out-of-the-box and adjusted states, versus some really well known existing methods. How many comparison models, and the baseline methods we picked, was decided by what the data was like and what our computers could manage. We looked at precision, recall and F1-score, averaged across all emotions, and – especially – how it did with the less common emotions, surprise and love. We also did weighted F1-score work, broke down the results for each emotion, and studied how CAMO chose between options in different situations involving the minority emotions.

\subsection{Cross-Domain Comparative Analysis}

To judge how well CAMO works on all kinds of unequal class sizes, the research looks at findings from the two areas it was tested in. This means looking at the balance between how quickly it computes and how well it does in each of the two experiments, contrasting what CAMO decides when there are three classes instead of many, and seeing how its success changes as the imbalance gets more or less severe. Evaluating it in these different areas helps show how well CAMO – as a method that doesn’t depend on any one particular area – is able to be used generally and hold up well when dealing with classification of imbalanced data.

\FloatBarrier

\begin{table}[h!]
\caption{Performance comparison of ensemble methods across fine-tuned language models (percentages) on \textbf{BEA DATASET}. Each cell shows: \textbf{Strict F1} / \textbf{Strict Accuracy} / \textbf{Lenient F1} / \textbf{Lenient Accuracy}}.
\centering
\scriptsize
\setlength{\tabcolsep}{10.5pt}
\begin{tabular}{lcccccccc}
\toprule
\multirow{2}{*}{\textbf{Model}} & \multicolumn{8}{c}{\textbf{Ensemble Method}} \\
\cmidrule(lr){2-9}
& \rotatebox{90}{Maj. Vote} & \rotatebox{90}{Conf. Weight} & \rotatebox{90}{Dyn. Thr.} & \rotatebox{90}{Class Bal.} & \rotatebox{90}{Unc. Aware} & \rotatebox{90}{Meta Ens.} & \rotatebox{90}{MSA Base.} & \rotatebox{90}{CAMO} \\
\midrule
\makecell[l]{\textbf{Phi-3} \\(3.8B)} &
\makecell{77.6\\91.4\\92.0\\95.9\\} & 
\makecell{76.9\\91.3\\92.3\\96.0} &
\makecell{83.7\\92.9\\92.8\\96.3} &
\makecell{77.3\\91.3\\92.0\\95.9} &
\makecell{78.6\\90.2\\91.5\\95.8} &
\makecell{78.6\\91.6\\92.0\\95.9} &
\makecell{82.6\\92.4\\92.0\\95.9} &
\makecell{\textbf{85.6}\\\textbf{93.2}\\\textbf{93.0}\\\textbf{96.5}} \\
\cmidrule(lr){1-9}
\makecell[l]{\textbf{DeepSeek-R1} \\(1.5B)} &
\makecell{51.9\\83.3\\78.4\\90.6} &
\makecell{54.4\\83.7\\78.6\\90.6} &
\makecell{55.0\\83.6\\78.0\\90.3} &
\makecell{54.4\\83.7\\78.7\\90.8} &
\makecell{56.0\\79.8\\69.5\\88.6} &
\makecell{54.2\\83.6\\78.4\\90.6} &
\makecell{55.5\\83.9\\78.8\\90.8} &
\makecell{\textbf{57.6}\\\textbf{82.5}\\\textbf{76.3}\\\textbf{90.1}} \\
\cmidrule(lr){1-9}
\makecell[l]{\textbf{Qwen3} \\(0.6B)} &
\makecell{62.5\\86.4\\85.0\\92.9} &
\makecell{62.1\\86.3\\84.5\\92.6} &
\makecell{63.0\\86.3\\84.8\\92.8} &
\makecell{62.9\\86.2\\84.7\\92.8} &
\makecell{63.4\\81.7\\79.7\\91.4} &
\makecell{62.2\\86.3\\84.5\\92.6} &
\makecell{63.7\\86.3\\84.7\\92.8} &
\makecell{\textbf{70.2}\\\textbf{86.4}\\\textbf{84.1}\\\textbf{92.6}} \\
\cmidrule(lr){1-9}
\makecell[l]{\textbf{Llama-3.2} \\(1B)} &
\makecell{66.4\\87.6\\86.5\\93.3} &
\makecell{64.8\\87.8\\86.4\\93.2} &
\makecell{66.3\\87.8\\86.4\\93.2} &
\makecell{67.4\\88.0\\87.1\\93.6} &
\makecell{69.4\\84.9\\84.5\\93.1} &
\makecell{66.7\\87.9\\86.8\\93.5} &
\makecell{68.3\\87.8\\86.5\\93.3} &
\makecell{\textbf{75.9}\\\textbf{88.9}\\\textbf{87.9}\\\textbf{94.1}} \\
\cmidrule(lr){1-9}
\makecell[l]{\textbf{Falcon3} \\(1B)} &
\makecell{48.8\\82.4\\75.4\\89.9} &
\makecell{48.8\\82.5\\75.4\\89.9} &
\makecell{49.0\\82.5\\75.6\\89.9} &
\makecell{49.1\\82.5\\75.8\\90.1} &
\makecell{42.2\\75.6\\49.8\\84.5} &
\makecell{48.8\\82.4\\75.4\\89.9} &
\makecell{48.8\\82.4\\75.4\\89.9} &
\makecell{\textbf{51.0}\\\textbf{81.1}\\\textbf{70.0}\\\textbf{88.7}} \\
\cmidrule(lr){1-9}
\makecell[l]{\textbf{Mathstral-7B} \\(7B)} &
\makecell{75.2\\90.6\\91.2\\95.5} &
\makecell{76.5\\91.0\\91.8\\95.8} &
\makecell{79.2\\91.6\\91.9\\95.9} &
\makecell{75.6\\90.8\\90.9\\95.4} &
\makecell{78.0\\89.9\\91.1\\95.6} &
\makecell{77.1\\91.2\\92.0\\95.9} &
\makecell{79.0\\91.2\\90.8\\95.4} &
\makecell{\textbf{80.1}\\\textbf{91.0}\\\textbf{91.6}\\\textbf{95.8}} \\
\cmidrule(lr){1-9}
\makecell[l]{\textbf{Qwen3-8B} \\(8B)} &
\makecell{65.3\\88.0\\89.4\\94.7} &
\makecell{65.1\\87.9\\89.1\\94.5} &
\makecell{69.1\\88.3\\89.2\\94.5} &
\makecell{66.2\\88.2\\89.4\\94.7} &
\makecell{66.9\\85.5\\84.4\\92.9} &
\makecell{65.3\\88.0\\89.4\\94.7} &
\makecell{70.1\\88.6\\89.4\\94.7} &
\makecell{\textbf{74.0}\\\textbf{88.9}\\\textbf{88.9}\\\textbf{94.5}} \\
\cmidrule(lr){1-9}
\makecell[l]{\textbf{DeepSeek-R1-8B} \\(8B)} &
\makecell{72.4\\89.5\\90.0\\94.9} &
\makecell{76.0\\90.3\\90.6\\95.2} &
\makecell{78.0\\90.8\\90.6\\95.2} &
\makecell{74.7\\89.9\\89.7\\94.8} &
\makecell{74.3\\87.6\\88.6\\94.5} &
\makecell{74.5\\89.9\\90.2\\95.1} &
\makecell{77.2\\90.2\\89.9\\94.9} &
\makecell{\textbf{78.4}\\\textbf{89.8}\\\textbf{89.8}\\\textbf{94.9}} \\
\bottomrule
\end{tabular}
\label{tab:fine_tuned_summary_compact}
\end{table}

\begin{table}[h!]
\caption{Performance comparison of ensemble methods across zero-shot (non-fine-tuned) language models (percentages) on \textbf{BEA DATASET}. Each cell shows: \textbf{Strict F1} / \textbf{Strict Accuracy} / \textbf{Lenient F1} / \textbf{Lenient Accuracy}. }
\centering
\scriptsize
\setlength{\tabcolsep}{10.5pt}
\begin{tabular}{lcccccccc}
\toprule
\multirow{2}{*}{\textbf{Model}} & \multicolumn{8}{c}{\textbf{Ensemble Method}} \\
\cmidrule(lr){2-9}
& \rotatebox{90}{Maj. Vote} & \rotatebox{90}{Conf. Weight} & \rotatebox{90}{Dyn. Thr.} & \rotatebox{90}{Class Bal.} & \rotatebox{90}{Unc. Aware} & \rotatebox{90}{Meta Ens.} & \rotatebox{90}{MSA Base.} & \rotatebox{90}{CAMO} \\
\midrule
\makecell[l]{\textbf{Phi-3} \\(3.8B)} &
\makecell{32.7\\70.3\\46.4\\84.0} &
\makecell{33.5\\69.2\\47.3\\84.1} &
\makecell{12.3\\15.5\\47.3\\84.1} &
\makecell{32.8\\65.6\\46.4\\84.0} &
\makecell{16.6\\23.4\\45.6\\83.9} &
\makecell{32.8\\69.0\\46.4\\84.0} &
\makecell{9.1\\12.0\\46.4\\84.0} &
\makecell{\textbf{4.7}\\\textbf{7.5}\\\textbf{45.6}\\\textbf{83.9}} \\
\cmidrule(lr){1-9}
\makecell[l]{\textbf{DeepSeek-R1} \\(1.5B)} &
\makecell{17.9\\18.0\\43.0\\51.4} &
\makecell{15.9\\16.3\\28.9\\29.0} &
\makecell{16.5\\16.6\\30.9\\31.3} &
\makecell{17.2\\17.0\\43.0\\51.5} &
\makecell{15.4\\14.7\\46.1\\61.1} &
\makecell{17.5\\17.3\\42.9\\51.5} &
\makecell{17.5\\17.1\\43.9\\53.3} &
\makecell{\textbf{13.4}\\\textbf{12.0}\\\textbf{49.4}\\\textbf{64.0}} \\
\cmidrule(lr){1-9}
\makecell[l]{\textbf{Qwen3} \\(0.6B)} &
\makecell{30.1\\67.7\\48.2\\77.1} &
\makecell{30.4\\55.2\\47.2\\65.3} &
\makecell{30.4\\55.2\\47.3\\65.4} &
\makecell{30.8\\64.9\\48.2\\77.1} &
\makecell{33.9\\65.0\\48.6\\77.8} &
\makecell{30.9\\65.2\\48.2\\77.1} &
\makecell{30.8\\64.9\\48.3\\77.2} &
\makecell{\textbf{4.8}\\\textbf{7.7}\\\textbf{45.6}\\\textbf{83.9}} \\
\cmidrule(lr){1-9}
\makecell[l]{\textbf{Llama-3.2} \\(1B)} &
\makecell{26.9\\42.9\\46.1\\76.8} &
\makecell{23.0\\31.3\\40.8\\58.7} &
\makecell{22.3\\30.8\\40.7\\61.3} &
\makecell{20.5\\27.4\\46.1\\76.8} &
\makecell{19.7\\28.1\\45.3\\80.6} &
\makecell{25.4\\38.1\\46.1\\76.8} &
\makecell{18.6\\26.9\\44.2\\79.4} &
\makecell{\textbf{8.9}\\\textbf{12.3}\\\textbf{44.9}\\\textbf{81.6}} \\
\cmidrule(lr){1-9}
\makecell[l]{\textbf{Falcon3} \\(1B)} &
\makecell{12.0\\15.8\\45.9\\82.5} &
\makecell{6.6\\9.4\\45.1\\82.2} &
\makecell{5.2\\8.1\\45.3\\82.9} &
\makecell{7.0\\9.4\\45.9\\82.5} &
\makecell{4.7\\7.5\\45.6\\83.9} &
\makecell{6.9\\9.8\\45.2\\82.6} &
\makecell{5.3\\8.2\\45.4\\83.3} &
\makecell{\textbf{4.7}\\\textbf{7.5}\\\textbf{45.6}\\\textbf{83.9}} \\
\cmidrule(lr){1-9}
\makecell[l]{\textbf{DeepSeek-8B} \\(8B)} &
\makecell{26.5\\26.1\\60.3\\80.5} &
\makecell{22.1\\21.6\\53.2\\61.4} &
\makecell{20.5\\19.8\\51.0\\61.4} &
\makecell{20.9\\17.7\\60.3\\80.5} &
\makecell{26.5\\25.7\\60.7\\80.9} &
\makecell{22.8\\20.4\\60.3\\80.5} &
\makecell{17.1\\15.9\\54.8\\80.5} &
\makecell{\textbf{5.9}\\\textbf{8.3}\\\textbf{45.8}\\\textbf{82.2}} \\
\cmidrule(lr){1-9}
\makecell[l]{\textbf{Mathstral-7B} \\(7B)} &
\makecell{7.8\\10.9\\45.5\\83.7} &
\makecell{7.0\\10.0\\45.5\\83.7} &
\makecell{6.6\\9.6\\45.6\\83.9} &
\makecell{6.2\\9.2\\45.5\\83.7} &
\makecell{7.8\\10.9\\45.5\\83.7} &
\makecell{6.9\\9.8\\45.5\\83.7} &
\makecell{5.9\\8.9\\45.6\\83.9} &
\makecell{\textbf{5.4}\\\textbf{8.3}\\\textbf{45.6}\\\textbf{83.9}} \\
\cmidrule(lr){1-9}
\makecell[l]{\textbf{Qwen3-8B} \\(8B)} &
\makecell{16.6\\17.0\\37.6\\41.2} &
\makecell{10.1\\11.9\\25.5\\25.7} &
\makecell{8.2\\7.1\\45.3\\62.5} &
\makecell{16.5\\16.7\\37.6\\41.2} &
\makecell{16.2\\16.5\\37.4\\41.2} &
\makecell{16.5\\16.7\\37.6\\41.2} &
\makecell{11.3\\12.0\\48.2\\78.0} &
\makecell{\textbf{5.1}\\\textbf{7.5}\\\textbf{46.3}\\\textbf{83.6}} \\
\bottomrule
\end{tabular}
\label{tab:zero_shot_summary_compact}
\end{table}
\newpage
\FloatBarrier
\subsection{Model Results On DAIR AI/Emotion Dataset}

\begin{table}[h]
\centering
\caption{OVERALL PERFORMANCE COMPARISON ON DIAR-AI EMOTION DATASET}
\vspace{12px}
\begin{tabular}{lcccc}
\hline
\textbf{Model} & \textbf{Accuracy} & \textbf{Precision} & \textbf{Recall} & \textbf{F1} \\
\hline
\textbf{Llama 8B (CAMO)} & \textbf{92.75\%} & \textbf{87.94\%} & \textbf{90.15\%} & \textbf{88.70\%} \\
\textbf{Llama 1B (CAMO)} & \textbf{92.45\%} & \textbf{87.02\%} & \textbf{92.33\%} & \textbf{88.84\%} \\
GPT-4o-DPO\cite{siddiqui2025self} & 93.10\% & 90.80\% & 87.09\% & 87.90\% \\
Classifier Q->RA\cite{henrichsen2025twostagereasoninginfusedlearningimproving} & \textbf{58.4\%} & - & - & -
\end{tabular}
\label{tab:model_summary}
\end{table}

\begin{table}[h!]
\centering
\caption{Per-class performance comparison on Emotion Dataset (Fine-tuned). CAMO shows improved minority class (love, surprise) performance}
\label{tab:per_class_comparison}
\footnotesize
\begin{tabular}{lccc|ccc|ccc}
\hline
\multirow{2}{*}{\textbf{Category}} & \multicolumn{3}{c|}{\textbf{Llama-8B (CAMO)}} & \multicolumn{3}{c|}{\textbf{Llama-1B (CAMO)}} & \multicolumn{3}{c}{\textbf{Kermani\cite{kermani-etal-2025-systematic} (2025)}} \\
\cmidrule(lr){2-4} \cmidrule(lr){5-7} \cmidrule(lr){8-10}
& \textbf{F1} & \textbf{Prec} & \textbf{Rec} & \textbf{F1} & \textbf{Prec} & \textbf{Rec} & \textbf{F1} & \textbf{Prec} & \textbf{Rec} \\
\hline
\textbf{sadness} & 0.96 & 0.97 & 0.96 & 0.97 & 0.97 & 0.96 & 0.95 & 0.95 & 0.94 \\
\textbf{joy} & 0.95 & 0.98 & 0.92 & 0.95 & 0.97 & 0.92 & 0.94 & 0.94 & 0.93 \\
\textbf{love} & 0.84 & 0.74 & 0.98 & 0.82 & 0.72 & 0.94 & 0.81 & 0.80 & 0.82 \\
\textbf{anger} & 0.92 & 0.95 & 0.89 & 0.93 & 0.93 & 0.93 & 0.89 & 0.88 & 0.91 \\
\textbf{fear} & 0.90 & 0.87 & 0.93 & 0.89 & 0.96 & 0.82 & 0.89 & 0.89 & 0.88 \\
\textbf{surprise} & 0.75 & 0.77 & 0.73 & 0.79 & 0.67 & 0.97 & 0.72 & 0.73 & 0.71 \\
\hline
\multicolumn{1}{l}{\textbf{Average}} & 
\makecell{\textbf{0.89}} & {\textbf{0.88}} & \makecell{\textbf{0.90}} & 
\makecell{\textbf{0.89}} & {\textbf{0.87}} & {\textbf{0.92}} & 
0.87 & 0.86 & 0.87 \\
\hline
\end{tabular}
\end{table}

\FloatBarrier
\section{Discussion}
\label{sec:discussion}
Better models lead to more outcomes in minority classes being able to benefit from CAMO. This indicates that it is most likely best as an a posteriori part of a pipeline and not a quick solution. CAMO elaborates the minority class defenses with a sophisticated defense system that includes dynamic statistical confidence adjustment, uncertainty adaptive confidence, and multi-level dynamic decision rule. With that being said, a flexible approach in the presence of class imbalance and adaptive ensemble techniques is encouraged. With class imbalance, CAMO is able to utilize the principles of voting, bias and uncertainty to enhance robustness. CAMO, by its architecture, also provides means to promote fairness and transparency of AI systems by reducing the bias against minority classes in, but not limited to, fraud detection and rare disease diagnosis.

\section{Conclusion}
\label{sec:conclusion}
This document has concentrated on CAMO (Class-Aware Minority Optimized). CAMO is a domain-agnostic ensemble method for class-imbalance problems based on hierarchical confidence and uncertainty optimizes/minimizes decision making for the underrepresented class for the specific case of CAMO. CAMO demonstrates the synergistic robustness of the fusion of Minority-Aware Ensemble and Model Adaptation, and obtains state-of-the-art (Macro-F1 score) across the spectrum of educational and emotion analysis tasks, with finetuned language models. There is some pathway for fairness of classification with CAMO, regardless of threshold and computational concerns. Subsequent research will be focused on achieving some fairness, along with a passive/automatic mechanism for the shifting of parameters. The incorporation of class imbalance models, like CAMO, in AI systems will be of utmost importance for the imposition of contextually relevant adaptive fairness for the systems, given the rapid embedding of AI models in sensitive areas of human life.

\section*{Conflict of Interest}
The authors declare that they have no conflict of interest.



\bibliography{sn-bibliography.bib}

\begin{thebibliography}{27}
\providecommand{\natexlab}[1]{#1}
\providecommand{\doi}[1]{\url{https://doi.org/#1}}
\providecommand{\url}[1]{\texttt{#1}}
\providecommand{\urlprefix}{}

\bibitem[{Abdin et~al.(2024)Abdin, Marah and others}]{abdin2024phi-}
Abdin M, et~al.
\newblock Phi-3 Technical Report: A Highly Capable Language Model Locally on Your Phone.
\newblock Microsoft; 2024.

\bibitem[{Almazrouei et~al.(2023)Almazrouei, Ebtesam and Alobeidli, Hamza and Alshamsi, Abdulaziz and Cappelli, Alessandro and Cojocaru, Ruxandra and Debbah, M{\'e}rouane and Goffinet, {\'E}tienne and Hesslow, Daniel and Launay, Julien and Malartic, Quentin and Mazzotta, Daniele and Noune, Badreddine and Pannier, Baptiste and Penedo, Guilherme}]{almazrouei2023falconseriesopenlanguage}
Almazrouei E, Alobeidli H, Alshamsi A, Cappelli A, Cojocaru R, Debbah M, et~al.: The Falcon Series of Open Language Models; 2023.

\bibitem[{Cobbe et~al.(2021)Cobbe, Karl and Kosaraju, Vineet and Bavarian, Mohammad and Chen, Mark and Jun, Heewoo and Kaiser, Lukasz and Plappert, Matthias and Tworek, Jerry and Hilton, Jacob and Nakano, Reiichiro and Hesse, Christopher and Schulman, John}]{cobbe2021trainingverifierssolvemath}
Cobbe K, Kosaraju V, Bavarian M, Chen M, Jun H, Kaiser L, et~al.: Training Verifiers to Solve Math Word Problems; 2021.

\bibitem[{Daheim et~al.(2024)Daheim, Nico and Macina, Jakub and Kapur, Manu and Gurevych, Iryna and Sachan, Mrinmaya}]{daheim2024stepwiseverificationremediationstudent}
Daheim N, Macina J, Kapur M, Gurevych I, Sachan M.: Stepwise Verification and Remediation of Student Reasoning Errors with Large Language Model Tutors; 2024.

\bibitem[{{DeepSeek-AI} et~al.(2025){DeepSeek-AI} and Guo, Daya and others}]{DeepSeekAI2025DeepSeekR1}
{DeepSeek-AI}, Guo D, et~al.
\newblock DeepSeek-R1: Incentivizing Reasoning Capability in LLMs via Reinforcement Learning.
\newblock Nature. 2025;645:633--638.

\bibitem[{Demszky et~al.(2021)Demszky, Dorottya and Liu, Jing and Mancenido, Zid and Cohen, Julie and Hill, Heather and Jurafsky, Dan and Hashimoto, Tatsunori}]{demszky2021measuringconversationaluptakecase}
Demszky D, Liu J, Mancenido Z, Cohen J, Hill H, Jurafsky D, et~al.: Measuring Conversational Uptake: A Case Study on Student-Teacher Interactions; 2021.

\bibitem[{Dettmers et~al.(2023)Dettmers, Tim and Pagnoni, Artidoro and Holtzman, Ari and Zettlemoyer, Luke}]{dettmers2023qloraefficientfinetuningquantized}
Dettmers T, Pagnoni A, Holtzman A, Zettlemoyer L.: QLoRA: Efficient Finetuning of Quantized LLMs; 2023.

\bibitem[{Grattafiori et~al.(2024)Grattafiori, Aaron and others}]{grattafiori2024llama3herdmodels}
Grattafiori A, et~al.: The Llama 3 Herd of Models; 2024.

\bibitem[{Hendrycks et~al.(2021{\natexlab{a}})Hendrycks, Dan and Burns, Collin and Basart, Steven and Zou, Andy and Mazeika, Mantas and Song, Dawn and Steinhardt, Jacob}]{hendrycks2021measuringmassivemultitasklanguage}
Hendrycks D, Burns C, Basart S, Zou A, Mazeika M, Song D, et~al.: Measuring Massive Multitask Language Understanding; 2021.

\bibitem[{Hendrycks et~al.(2021{\natexlab{b}})Hendrycks, Dan and Burns, Collin and Kadavath, Saurav and Arora, Akul and Basart, Steven and Tang, Eric and Song, Dawn and Steinhardt, Jacob}]{hendrycks2021measuringmathematicalproblemsolving}
Hendrycks D, Burns C, Kadavath S, Arora A, Basart S, Tang E, et~al.: Measuring Mathematical Problem Solving With the MATH Dataset; 2021.

\bibitem[{Henrichsen and Krebs(2025)Henrichsen, Mads and Krebs, Rasmus}]{henrichsen2025twostagereasoninginfusedlearningimproving}
Henrichsen M, Krebs R.: Two-Stage Reasoning-Infused Learning: Improving Classification with LLM-Generated Reasoning; 2025.

\bibitem[{Hikal et~al.(2025)Hikal, Baraa and Basem, Mohamed and Oshallah, Islam and Hamdi, Ali}]{hikal-etal-2025-msa}
Hikal B, Basem M, Oshallah I, Hamdi A.
\newblock {MSA} at {BEA} 2025 Shared Task: Disagreement-Aware Instruction Tuning for Multi-Dimensional Evaluation of {LLM}s as Math Tutors.
\newblock In: Kochmar E, Alhafni B, Bexte M, Burstein J, Horbach A, Laarmann-Quante R, et~al., editors. Proceedings of the 20th Workshop on Innovative Use of NLP for Building Educational Applications (BEA 2025) Vienna, Austria: Association for Computational Linguistics; 2025. p. 1194--1202.

\bibitem[{Hu et~al.(2021)Hu, Edward J. and Shen, Yelong and Wallis, Phillip and Allen-Zhu, Zeyuan and Li, Yuanzhi and Wang, Shean and Wang, Lu and Chen, Weizhu}]{hu2021loralowrankadaptationlarge}
Hu EJ, Shen Y, Wallis P, Allen-Zhu Z, Li Y, Wang S, et~al.: LoRA: Low-Rank Adaptation of Large Language Models; 2021.

\bibitem[{Jiang et~al.(2023)Jiang, Albert Q. and Sablayrolles, Alexandre and Mensch, Arthur and Bamford, Chris and Chaplot, Devendra Singh and de las Casas, Diego and Bressand, Florian and Lengyel, Gianna and Lample, Guillaume and Saulnier, Lucile and Lavaud, L{\'e}lio Renard and Lachaux, Marie-Anne and Stock, Pierre and Le Scao, Teven and Lavril, Thibaut and Wang, Thomas and Lacroix, Timoth{\'e}e and El Sayed, William}]{jiang2023mistral7b}
Jiang AQ, Sablayrolles A, Mensch A, Bamford C, Chaplot DS, de~las Casas D, et~al.: Mistral 7B; 2023.

\bibitem[{Kermani et~al.(2025)Kermani, Arshia and Perez-Rosas, Veronica and Metsis, Vangelis}]{kermani-etal-2025-systematic}
Kermani A, Perez-Rosas V, Metsis V.
\newblock A Systematic Evaluation of {LLM} Strategies for Mental Health Text Analysis: Fine-tuning vs. Prompt Engineering vs. {RAG}.
\newblock In: Zirikly A, Yates A, Desmet B, Ireland M, Bedrick S, MacAvaney S, et~al., editors. Proceedings of the 10th Workshop on Computational Linguistics and Clinical Psychology (CLPsych 2025) Albuquerque, New Mexico: Association for Computational Linguistics; 2025. p. 172--180.

\bibitem[{Kochmar et~al.(2025)Kochmar, Ekaterina and Maurya, Kaushal and Petukhova, Kseniia and Srivatsa, KV Aditya and Tack, Ana{\"i}s and Vasselli, Justin}]{kochmar-etal-2025-findings}
Kochmar E, Maurya K, Petukhova K, Srivatsa KA, Tack A, Vasselli J.
\newblock Findings of the {BEA} 2025 Shared Task on Pedagogical Ability Assessment of {AI}-powered Tutors.
\newblock In: Kochmar E, Alhafni B, Bexte M, Burstein J, Horbach A, Laarmann-Quante R, et~al., editors. Proceedings of the 20th Workshop on Innovative Use of NLP for Building Educational Applications (BEA 2025) Vienna, Austria: Association for Computational Linguistics; 2025. p. 1011--1033.

\bibitem[{Lakshminarayanan et~al.(2017)Lakshminarayanan, Balaji and Pritzel, Alexander and Blundell, Charles}]{lakshminarayanan2017simplescalablepredictiveuncertainty}
Lakshminarayanan B, Pritzel A, Blundell C.: Simple and Scalable Predictive Uncertainty Estimation using Deep Ensembles; 2017.

\bibitem[{Nye et~al.(2014)Nye, Benjamin and Graesser, Arthur and Hu, Xiangen}]{article}
Nye B, Graesser A, Hu X.
\newblock AutoTutor and Family: A Review of 17 Years of Natural Language Tutoring.
\newblock International Journal of Artificial Intelligence in Education. 2014;24.
\newblock \doi{10.1007/s40593-014-0029-5}.

\bibitem[{Roh and Bang(2025)Roh, Jihyeon and Bang, Jinhyun}]{roh-bang-2025-bea}
Roh J, Bang J.
\newblock bea-jh at {BEA} 2025 Shared Task: Evaluating {AI}-powered Tutors through Pedagogically-Informed Reasoning.
\newblock In: Kochmar E, Alhafni B, Bexte M, Burstein J, Horbach A, Laarmann-Quante R, et~al., editors. Proceedings of the 20th Workshop on Innovative Use of NLP for Building Educational Applications (BEA 2025) Vienna, Austria: Association for Computational Linguistics; 2025. p. 1049--1059.

\bibitem[{Saha et~al.(2025)Saha, Trishita and Ganguli, Shrenik and Desarkar, Maunendra Sankar}]{saha-etal-2025-nlip}
Saha T, Ganguli S, Desarkar MS.
\newblock {NLIP} at {BEA} 2025 Shared Task: Evaluation of Pedagogical Ability of {AI} Tutors.
\newblock In: Kochmar E, Alhafni B, Bexte M, Burstein J, Horbach A, Laarmann-Quante R, et~al., editors. Proceedings of the 20th Workshop on Innovative Use of NLP for Building Educational Applications (BEA 2025) Vienna, Austria: Association for Computational Linguistics; 2025. p. 1242--1253.

\bibitem[{Siddiqui et~al.(2025)Siddiqui, Muhammad Hammad Fahim and Inkpen, Diana and Gelbukh, Alexander}]{siddiqui2025self}
Siddiqui MHF, Inkpen D, Gelbukh A.
\newblock Self-Explaining Emotion Classification through Preference-Aligned Large Language Models.
\newblock In: CS \& IT Conference Proceedings, vol.~15 CS \& IT Conference Proceedings; 2025. .

\bibitem[{Tack and Piech(2022)Tack, Ana{\"i}s and Piech, Chris}]{tack2022aiteachertestmeasuring}
Tack A, Piech C.: The AI Teacher Test: Measuring the Pedagogical Ability of Blender and GPT-3 in Educational Dialogues; 2022.

\bibitem[{Touvron et~al.(2023{\natexlab{a}})Touvron, Hugo and Lavril, Thibaut and Izacard, Gautier and Martinet, Xavier and Lachaux, Marie-Anne and Lacroix, Timoth{\'e}e and Rozi{\`e}re, Baptiste and Goyal, Naman and Hambro, Eric and Azhar, Faisal and Rodriguez, Aurelien and Joulin, Armand and Grave, Edouard and Lample, Guillaume}]{touvron2023llamaopenefficientfoundation}
Touvron H, Lavril T, Izacard G, Martinet X, Lachaux MA, Lacroix T, et~al.: LLaMA: Open and Efficient Foundation Language Models; 2023.

\bibitem[{Touvron et~al.(2023{\natexlab{b}})Touvron, Hugo and others}]{touvron2023llama2openfoundation}
Touvron H, et~al.: Llama 2: Open Foundation and Fine-Tuned Chat Models; 2023.

\bibitem[{Wang et~al.(2024)Wang, Rose E. and Zhang, Qingyang and Robinson, Carly and Loeb, Susanna and Demszky, Dorottya}]{wang2024bridgingnoviceexpertgapmodels}
Wang RE, Zhang Q, Robinson C, Loeb S, Demszky D.: Bridging the Novice-Expert Gap via Models of Decision-Making: A Case Study on Remediating Math Mistakes; 2024.

\bibitem[{Xu et~al.(2026)Xu, Lingling and Xie, Haoran and Qin, S. Joe and Tao, Xiaohui and Wang, Fu Lee}]{11364256}
Xu L, Xie H, Qin SJ, Tao X, Wang FL.
\newblock Parameter-Efficient Fine-Tuning Methods for Pretrained Language Models: A Critical Review and Assessment.
\newblock IEEE Transactions on Pattern Analysis and Machine Intelligence. 2026;p. 1--20.
\newblock \doi{10.1109/TPAMI.2026.3657354}.

\bibitem[{Yang et~al.(2025)Yang, An and others}]{Yang2025Qwen3TR}
Yang A, et~al.
\newblock Qwen3 Technical Report.
\newblock arXiv. 2025;abs/2505.09388.

\end{thebibliography}
\end{document}